\documentclass[10pt,twocolumn,letterpaper]{article}
\pdfoutput=1
\usepackage{wacv}
\usepackage{times}
\usepackage{epsfig}
\usepackage{graphicx}
\usepackage{amsmath}
\usepackage{amssymb}
\usepackage[svgnames]{xcolor}
\usepackage{pdfpages}

\wacvfinalcopy 

\def\wacvPaperID{****} 

\ifwacvfinal\fi
\begin{document}

\title{Semantic Matching by Weakly Supervised 2D Point Set Registration}

\author{Zakaria Laskar \\
Aalto University\\
{\tt\small zakaria.laskar@aalto.fi}
\and
Hamed R. Tavakoli \\
Aalto University\\
{\tt\small hamed.r-tavakoli@aalto.fi}
\and
Juho Kannala \\
Aalto University\\
{\tt\small juho.kannala@aalto.fi}
}

\maketitle
\ifwacvfinal\thispagestyle{empty}\fi


\begin{abstract}

   In this paper we address the problem of establishing correspondences between different instances of the same object. The problem is posed as finding the geometric transformation that aligns a given image pair. We use a convolutional neural network (CNN) to directly regress the parameters of the transformation model. The alignment problem is defined in the setting where an unordered set of semantic key-points per image are available, but, without the correspondence information. To this end we propose a novel loss function based on cyclic consistency that solves this 2D point set registration problem by inferring the optimal geometric transformation model parameters. We train and test our approach on a standard benchmark dataset Proposal-Flow (PF-PASCAL)\cite{proposal_flow}. The proposed approach achieves state-of-the-art results demonstrating the effectiveness of the method. In addition, we show our approach further benefits from additional training samples in PF-PASCAL generated by using category level information.
   
    
   
\end{abstract}

\section{Introduction}
\label{sec:intro}
Establishing correspondences between images has been a fundamental problem in computer vision for decades. Areas of application includes structure from motion, tracking, stereo fusion, and, optical flow among many others. Although challenging, these areas of research often deal with images of the same object or scene. However, the task of semantic matching are not based on similar assumptions and offers much more challenging variations in terms of appearance and geometry.

In this paper, we aim to tackle the problem of semantic matching using CNNs. Until recently, methods \cite{scnet_2},\cite{scnet_17} that use traditional hand-crafted descriptors \cite{SIFT},\cite{DAISY},\cite{HOG} have topped the performance tables. Similar to the success of deep-learning in other fields of computer vision \cite{imagenet}, CNNs have also made an impact in the field of semantic matching \cite{scnet},\cite{ignacio_cvpr'18},\cite{ignacio_cvpr'17}. However, this also comes with some challenges. Semantic matching is evaluated by measuring the pixel transfer error. To train a neural network for this task in a supervised manner, one of the following forms of supervision is needed: \emph{i)} geometric transformations like affine, thin-plate spline, homography or relative camera position with depth, and, \emph{ii)} flow fields that contain pixel level correspondence. However, one needs a large amount of image pairs with such supervision to train CNNs. Unlike optical flow \cite{flownet}, such amount of data is not available for the particular task of semantic matching. As such existing methods either use self-supervision \cite{ignacio_cvpr'17} or weak-supervision \cite{ignacio_cvpr'18}. On the other hand, obtaining images with sparse ground-truth semantic key-points along with their respective object categories is relatively simple for small datasets (e.g. Proposal Flow). Thereby, using pairwise combinations within a given object category, large number of image pairs can be obtained. However, without the correspondence information between their respective semantic key-points, it is not clear how to effectively utilize this form of supervision to learn a better semantic correspondence function.

In this paper, we cast the task of semantic matching as solving a 2D point set registration problem. Our goal is to infer the parameters of a transformation model that best aligns the key-points from each image in a given image pair. The only requirement is that the semantic key-point sets should have atleast partial overlap. This indirectly removes any assumptions that the point sets should have the same order or size. We use a CNN \cite{ignacio_cvpr'17} to predict the transformation parameters. In order to train the neural network on the end objective of key-point alignment, we propose a novel loss function based on nearest-neighbor cyclic consistency. The proposed loss is a function of the predicted transformation parameters and thus allows back-propagation to train the parameters of the CNN. Given a source-target image pair, the key-points from source image are projected onto the target image using the estimated geometric transformation parameters. Thereafter, each projected source point is assigned a nearest neighbor from the ground-truth target key-points and the Euclidean distance between them constitutes the nearest-neighbor constraint. In addition, we impose the cycle consistency constraint that ensures the projected source points re-project close to the original source points under backward transformation. Results show that the proposed method significantly outperforms the baseline CNN geometric model \cite{ignacio_cvpr'17} on semantic matching datasets. We also analyze and demonstrate that the combination of two constraints is particularly important in achieving better performance.

\section{Related Work}
 
\textbf{Semantic matching.} 
Like other fields of computer vision, SIFT \cite{SIFT} features and descriptors have been the traditional choice in the field of semantic matching. SIFTFlow \cite{SIFTflow} computes dense SIFT features, followed by hierarchical optimization of matching cost to obtain dense pixel level flow. Yang \etal \cite{yang2014daisy} instead use DAISY \cite{DAISY} descriptors. While these approaches use descriptors at pixel level for matching, Ham \etal \cite{proposal_flow} introduced \textit{proposal flow} that uses region proposals as matching elements. They use HOG descriptors to match region proposals. Similarly, Taniai \etal \cite{taniai} also use HOG descriptors to jointly perform the task of co-segmentation and generate correspondence flow field.

Using CNN descriptors from networks pre-trained on ImageNet \cite{imagenet} instead of traditional hand-engineered descriptors have shown promising results \cite{Babenko_2015_ICCV}. However, \cite{proposal_flow} shows that the descriptors do not generalize well to the domain of semantic matching. On the other hand, optical flow based methods \cite{flownet} demonstrated that when trained end-to-end, CNNs can outperform hand-engineered descriptors in obtaining dense correspondence. As such, \cite{scnet}, and, \cite{fcss}, fine-tune CNN parameters by computing the loss in the \textit{proposal flow} framework. Similarly, Choy \etal \cite{UCN} propose a universal correspondence network to learn dense CNN descriptors using metric learning. The final correspondence flow field is obtained by matching putative regions using the CNN descriptors. Instead, \cite{zhou_cycle}, \cite{ignacio_cvpr'18}, \cite{ignacio_cvpr'17} directly output the correspondence map. \cite{zhou_cycle} directly outputs dense optical flow styled correspondence field, while, \cite{ignacio_cvpr'18},\cite{ignacio_cvpr'17} outputs the parameters of a transformation model. The geometric transformation model is then used to generate correspondence flow field.

As generating dense ground-truth correspondence field is a challenging task, recent methods have shown that photometric consistency can be used to train CNNs to predict correspondence flow. In particular, Zhou \etal \cite{zhou_ego} uses it to predict relative camera motion and depth for a given image pair. \cite{semi_super_GAN_flow} uses the photometric loss in a semi-supervised manner with GAN (Generative Adversarial Network) to perform domain adaptation from synthetic to real optical flow datasets. In the field of semantic matching, color (or photometric) constancy constraint is not valid due to the appearance variation between different instances of similar objects.


\begin{figure*}[t!]
		\centering
        \includegraphics[width=0.75\textwidth]{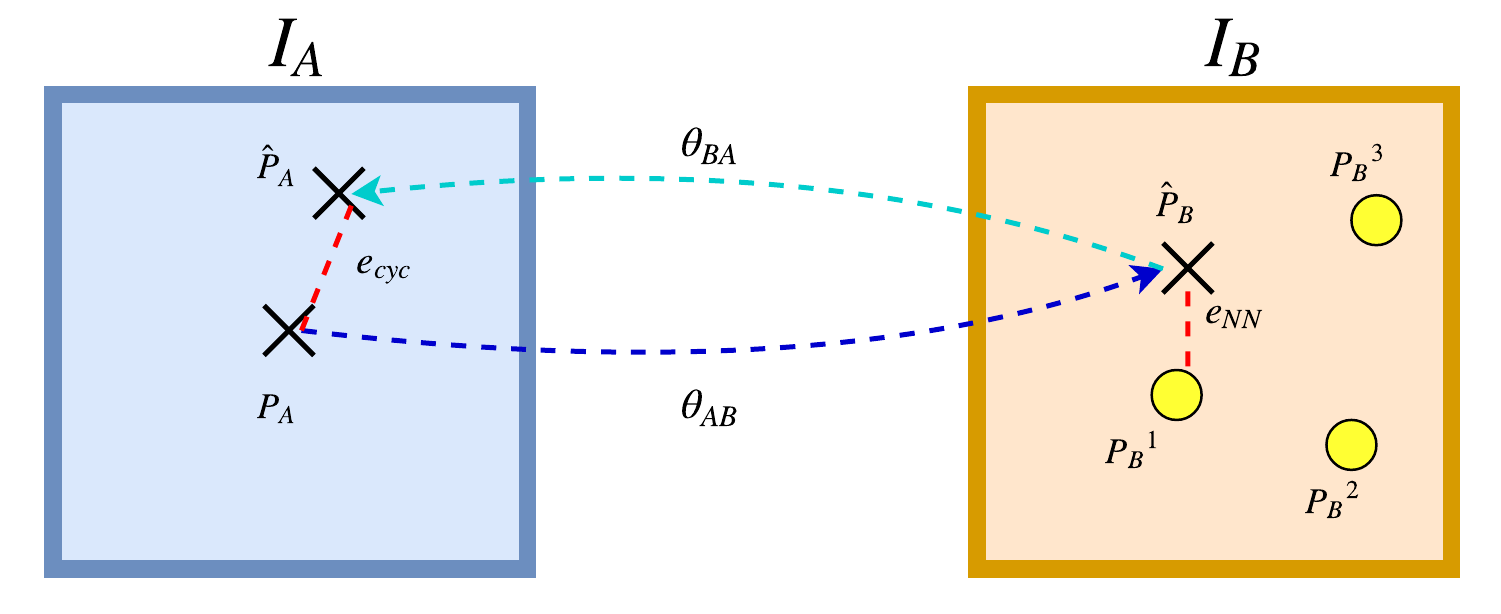}
        \caption{\textbf{Nearest Neighbor Cycle Consistency.}. The Figure shows the nearest neighbor and the cyclic consistency constraint that is used to train a geometric transformation network. $e_{cyc}$ represents the cyclic consistency loss which is the re-projection error of $P_A$ under the transformations $\theta_{AB}$ o $\theta_{BA}$. $e_{NN}$ measure the error between the projected source point, $\hat{P}_A$ and its nearest neighbor target point (${P_B}^1$ in the Figure.). \label{cycle}}
\end{figure*}        

In \cite{ignacio_cvpr'17}, Rocco \etal generate image pairs by synthetically transforming a set of images. This is used to train the network for predicting the parameters of the transformation model. This was later extended to \cite{ignacio_cvpr'18}, where the same geometric model was trained on real image pairs in a weakly supervised manner. The weak supervision was in the form of image level correspondence and the loss was computed by feature correlation at positions consistent with the predicted transformation parameters.


\noindent\textbf{Cycle consistency and dual learning.}

The cycle consistency helps learning correspondent representations such that a mapping from a source space to a target space should result in a similar target to source representation once inputted to an invert mapping. To this end, the pairwise distances of such representations is minimized within source and target spaces via a distance function, creating a cycle, please see Figure~\ref{cycle}.

Our approach is inspired by the work of Zhou \etal~\cite{zhou_iccv2015} in which authors propose learning dense correspondences between images and their 3D CAD model by imposing a cycle-consistency condition. Zhou \etal\cite{zhou_cycle} extends the idea to the framework of CNNs by leveraging 3D models to create cyclic graphs between the rendered synthetic views and pairs of images. The network is made to predict transformations for image-image and image-synthetic pairs. Using the 4-cycle constraint, the synthetic-synthetic transformation is estimated and compared with the ground-truth to generate gradients. However, the method necessitates the availability of 3D models and sampling appropriate synthetic views.  

In this work, we take the same idea and employ it to key-point correspondences. That is minimizing the distance between the key-points in source image and its estimated position obtained by traversing a cycle using respective flow fields.

\section{Proposed Approach}
In this section, we first describe the framework and the desired properties of the geometric alignment model. Then we proceed to define various learning objectives for training a CNN to predict the parameters of the transformation model.

Correspondences can be established between a given source-target image pair by either predicting a dense optical flow \cite{flownet},\cite{zhou_cycle} or by regressing the parameters of a geometric model \cite{ignacio_cvpr'18},\cite{ignacio_cvpr'17}. The main objective is to obtain a mapping from source to target pixels and measuring the pixel transfer error. In our framework this pixel transfer error is computed at training and evaluation stages at only semantically meaningful pixel locations (key-points). However, these semantic key-points have structured local and global properties which can be better modelled by a geometric transformation model. This makes a geometric model more suitable than dense flow based models. As the error is measured at pixel level (i.e. at semantic key-point locations), the only requirement is that the geometric model should be differentiable. This allows computation of the gradient of the key-point transfer error w.r.t parameters of the geometric model. These gradients can then be back-propagated to the network parameters.

We therefore use the geometric CNN \cite{ignacio_cvpr'17} as our base network which uses synthetic transformations as ground-truth to learn a geometric mapping. The network outputs affine and thin-plate spline (tps) transformation parameters in an iterative manner. The network is trained using a grid loss as follows: A fixed grid of points $\mathbb{G} = \{g_i\}$, where $g \in \mathbb{R}^2$ and $Z = |G|$, is defined on the source image. Then using the estimated and ground truth synthetic transformations parameters, $\hat{\theta}$ and $\theta$, respective transformations $\mathbb{T}_{\hat{\theta}}$ and $\mathbb{T}_{\theta}$ are obtained. These transformations are then used to warp $G$. The grid loss is then computed in the space of warped grid locations : 

\begin{equation}
L_{ss} = \frac{1}{Z}\sum_{i=1}^{Z}{||\mathbb{T}_{\hat{\theta}}(G) - \mathbb{T}_{\theta}(G)||_2}
\label{self-super}
\end{equation}

\subsection{Inference}
 In the current setting, we replace the uniform grid locations $\mathbb{G}$ with the ground-truth semantic key-points $\mathbb{P}$. For a given source-target image pair $A,B$ we have as ground-truth a set of semantic key-points $\mathbb{P}_A$, $\mathbb{P}_B$ with cardinalities $|\mathbb{P}_A|=M$ and $ |\mathbb{P}_B|=N$. We consider the case, where the sets are unordered and having unequal cardinalities, i.e. $M \neq N$. Hence we define a correspondence map $C : \mathbb{R}^{M\times N} \Rightarrow \{0,1\}$, such that $C[a,b] = 1$ if $p_a \in \mathbb{P}_A$ and $p_b \in \mathbb{P}_B$ are in correspondence, else the elements of $C$ are 0. Therefore, equation \ref{self-super} can be re-written as 

\begin{equation}
L_{s} = \frac{1}{M}\sum_{\substack{i=1 \\ p_a,p_b|C[a,b]=1}}^{M}{||\mathbb{T}_{\hat{\theta}_{AB}}(p_a) - p_b||_2}
\label{super}
\end{equation}

The key difference with Equation \ref{self-super} is that there may not exist a single true global optimum that best aligns the key-points of a given image pair. However, observing a diverse set of image pairs from a given object category and corresponding key-points should make the network converge to the least-squares estimate.

\subsection{Learning objectives}
\label{leobj}
As defined in our problem setting, we only have knowledge of the semantic key-points $\mathbb{P}_A$, $\mathbb{P}_B$ and not the correspondence map, $C$. Thereby, in this section we propose a list of candidate loss functions to solve Equation \ref{super} without the correspondence information.

\medskip 
\noindent\textbf{Nearest-Neighbor}
Based on the principle of Iterative Closest Point(ICP) algorithm, the projected source points under the forward transformation are assigned the nearest-neighbor target points in the Euclidean space as correspondence. Thereafter, the error is computed as the Euclidean distance between each projected source points and its nearest neighbor target point.

\begin{equation}
L^{F}_{nn} = \frac{1}{M}\sum_{i=1}^{M}{\min_{p_b \in \mathbb{P}_B}||\mathbb{T}_{\hat{\theta}_{AB}}(p_a) - p_b||_2}
\label{fwd-nn}
\end{equation}



\noindent\textbf{Chamfer Distance}
Chamfer distance (CD) also works on the principle of minimizing distance between nearest neighbors for each point in a point set pair. In an earlier work, it has been used for 3D point cloud registration \cite{fan2017point}. In addition to Equation \ref{fwd-nn}, CD additionally measures the distance between each target point and its nearest-neighbor projected source point.

\begin{equation}
\begin{aligned}
L^{F}_{cd} = L^F_{nn}
+ \frac{1}{N}\sum_{i=1}^{N}{\min_{p_a \in \mathbb{P}_A}||\mathbb{T}_{\hat{\theta}_{AB}}(p_a) - p_b||_2}
\end{aligned}
\label{fwd-CD}
\end{equation}

\noindent\textbf{Cyclic Consistency}

While the nearest-neighbor policy works quite well in practice, it can fail under viewpoint change (c.f. Section \ref{ablation}). For instance a geometric transformation network pre-trained on simple synthetic affine transformations may not generalize well to real world samples with significant viewpoint variation. This will result in source points not being projected close to the correct target points. Thereby, in addition to the nearest-neighbor constraint, we also constraint the projected source points to re-project back to the original source points under the backward transformation. If the nearest-neighbor target point is assigned incorrectly to a source point, then the backward transformation will restrict the convergence of the points under the nearest-neighbor constraint. Thus the network has to search the space of transformation parameters such that a source points is projected close to a target point which re-projects back to the original source point under the backward transformation 
. This also highlights the importance of applying a geometric model to parameterize the transformation as it indirectly uses the global consensus between the semantic key-points in an image to solve the alignment problem.  The cyclic consistency loss can be combined with both the nearest-neighbor or the chamfer distance loss functions. For brevity, we only express in mathematical form the combination of cyclic consistency and nearest neighbor:

\begin{equation}
\begin{aligned}
L^{F}_{\small{nn-Cyc}} =  L^F_{nn}
+\frac{1}{M}\sum_{i=1}^{M}{||\mathbb{T}_{\hat{\theta}_{BA}}(\mathbb{T}_{\hat{\theta}_{AB}}(p_a)) - p_a||_2}
\end{aligned}
\label{cyclic-fwd-nn}
\end{equation}

The superscript $F$ denotes that the above loss functions are computed in the forward direction involving the source points only. The same can be computed in the backward direction for target key-points. Although backward transformation is computed in cyclic consistency, the loss is measured in the space of source points only and not the target points. Equations \ref{fwd-nn},\ref{fwd-CD}, and \ref{cyclic-fwd-nn} can be written in the backward direction as : 

\begin{equation}
L^B_{nn} = \frac{1}{N}\sum_{i=1}^{N}{\min_{p_a \in \mathbb{P}_A}||\mathbb{T}_{\hat{\theta}_{BA}}(p_b) - p_a||_2}
\label{bck-nn}
\end{equation}

\begin{equation}
\begin{aligned}
L^B_{cd} = L^B_{nn}
+ \frac{1}{N}\sum_{i=1}^{N}{\min_{p_b \in \mathbb{P}_B}||\mathbb{T}_{\hat{\theta}_{BA}}(p_b) - p_a||_2}
\end{aligned}
\label{bck-CD}
\end{equation}

\begin{equation}
\begin{aligned}
L^{B}_{\small{nn-Cyc}} = L^B_{nn}
+\frac{1}{N}\sum_{i=1}^{N}{||\mathbb{T}_{\hat{\theta}_{AB}}(\mathbb{T}_{\hat{\theta}_{BA}}(p_b)) - p_b||_2}
\end{aligned}
\label{cyclic-bck-nn}
\end{equation}

Therefore, the nearest-neighbor, CD, and nearest-neighbor cyclic consistency loss functions can be respectively defined as : 

\begin{equation}
\begin{aligned}
{L}_{\small{nn}} = L^{F}_{\small{nn}} + L^{B}_{\small{nn}}
\end{aligned}
\label{nn}
\end{equation}

\begin{equation}
\begin{aligned}
{L}_{\small{cd}} = L^{F}_{\small{cd}} + L^{B}_{\small{cd}}
\end{aligned}
\label{CD}
\end{equation}

\begin{equation}
\begin{aligned}
{L}_{\small{nn-Cyc}} = L^{F}_{\small{nn-Cyc}} + L^{B}_{\small{nn-Cyc}}
\end{aligned}
\label{cyclic-nn}
\end{equation}




\section{Experimental Results}
In this section we present the experimental settings to test the proposed method. 

\begin{table*}[]
\centering
\setlength\tabcolsep{1.5pt}
\scalebox{0.9}{
\begin{tabular}{|lllllllllllllllllllll|l|}
\hline
                  & aero  & bike  & bird  & boat  & bottle & bus   & car   & cat   & chair & cow   & table & dog   & horse & mbike & person & plant & sheep & sofa  & train & tv    & mean  \\ \hline \hline
LOM\cite{proposal_flow}        & 73.3  & 74.4  & 54.4  & 50.9  & 49.6   & 73.8  & 72.9  & 63.6  & 46.1  & 79.8  & 42.5         & 48.0    & 68.3  & 66.3      & 42.1   & 62.1        & 65.2  & 57.1  & 64.4  & 58.0    & 62.5  \\ 
SCNet-A           & 67.6  & 72.9  & 69.3  & 59.7  & 74.5   & 72.7  & 73.2  & 59.5  & 51.4  & 78.2  & 39.4         & 50.1  & 67.0    & 62.1      & 69.3   & 68.5        & 78.2  & 63.3  & 57.7  & 59.8  & 66.3  \\ 
SCNet-AG          & 83.9  & 81.4  & 70.6  & 62.5  & 60.6   & 81.3  & 81.2  & 59.5  & 53.1  & 81.2  & 62.0           & 58.7  & 65.5  & 73.3      & 51.2   & 58.3        & 60.0    & 69.3  & 61.5  & 80.0    & 69.7  \\ 
SCNet-AG+         & 85.5  & 84.4  & 66.3  & 70.8  & 57.4   & 82.7  & 82.3  & 71.6  & 54.3  & 95.8  & 55.2         & 59.5  & 68.6  & 75.0        & 56.3   & 60.4        & 60.0    & 73.7  & 66.5  & 76.7  & 72.2  \\ 
CNNGeo  & 82.4  & 80.9  & 85.9  & 47.2  & 57.8   & 83.1  & 92.8  & 86.9  & 43.8  & 91.7  & 28.1         & 76.4  & 70.2  & 76.6      & 68.9   & 65.7        & 80.0    & 50.1  & 46.3  & 60.6  & 71.9  \\ 
CNNGeo2  & 83.7  & 88.0    & 83.4  & 58.3  & 68.8   & 90.3  & 92.3  & 83.7  & 47.4  & 91.7  & 28.1         & 76.3  & 77.0    & 76.0        & 71.4   & 76.2        & 80    & 59.5  & 62.3  & 63.9  & 75.8  \\ 
CNNGeo-NN      & 86.1 & 87.1 & 79.7 & 70.8 & 70.3    & 98.1  & 93.0 & 74.2 & 54.5    & 91.7  & 32.8        & 70.3 & 66.2 & 76.2     & 69.4  & 65.2          & 80.0   & 50.3  & 77.8 & 78.9 & 76.7 \\ 
CNNGeo-CD          & 86.2 & 88.7 & 80.9 & 72.2 & 79.7  & 96.9 & 94.6 & 80.4 & 67.6 & 87.5  & 57.81       & 71.1 & 79.9 & 82.5     & 74.3  & 89.5       & 20.0    & 69.1 & 73.8 & \textbf{85.6} & 81.5 \\ 
CNNGeo-CD-Cyc & 87.1 & 89.0 & 84.3 & \textbf{76.4} & 85.9   & \textbf{98.1} & \textbf{95.6}  & 80.4 & \textbf{72.4} & 89.6 & \textbf{67.7}           & 80.7 & \textbf{89.2} & 83.9     & 76.1  & \textbf{94.3 }       & \textbf{100.0}   & \textbf{79.3} & 83.8    & \textbf{85.6} & 85.3 \\
CNNGeo-NN-Cyc & \textbf{89.9} & \textbf{90.2} & \textbf{88.6} & 70.8 & \textbf{89.1}   & 97.5 & 94.9  & \textbf{83.1} & 68.8 & \textbf{93.8} & 64.1          & \textbf{83.1} & 83.5 & \textbf{85.3}     & \textbf{77.7}  & 87.1        & 80.0   & 77.9 & \textbf{87.0}    & 83.9 & \textbf{85.7} \\ \hline
\end{tabular}}

\caption{\textbf{Per class PCK on PF-PASCAL dataset}. PCK threshold, $\alpha$ = 0.1. The proposed model outperforms the existing methods. However, our model uses more supervisory data than the current state-of-the-art CNNGeo2. The supervisory signal used here is the information of ground-truth semantic key-points, which is cheaper to obtain for a small number of samples like the Proposal Flow. Our proposed methods show that the baseline model CNNGeo can be extended to learn from weak key-point supervision resulting in much better semantic matching performance.\label{tab:results}}
\end{table*}


\subsection{Dataset}
\label{sec:dataset}
The proposed approach is trained and evaluated on PF-PASCAL dataset. First proposed in \cite{proposal_flow}, the dataset has since been a standard benchmark on variety of tasks related to learning the semantics of an image using deep learning \cite{ignacio_cvpr'18},\cite{scnet},\cite{ignacio_cvpr'17}. The dataset consists of 1400 image pairs selected from PASCAL-VOC \cite{proposal_flow} dataset. The image pairs come from 20 object categories and are annotated with corresponding key-point locations.

The split proposed in earlier works \cite{proposal_flow},\cite{ignacio_cvpr'18},\cite{ignacio_cvpr'17} is used to generate training, validation and test image pairs. This results in about 700,300 and 300 image pairs respectively. As deep learning models perform better with variations in training set, additional training samples are generated by random flipping of image pairs in training set. Although the correspondence information is available, under the given framework we do not use it in our training. We refer to this dataset as $D_l$, where $|D_l| \approx $ 2500.

In addition to random flipping, using the object category information, more number of image pairs can be generated. This is done by category specific pairwise combinations of images from different labeled image pairs. Due to the imbalance in the number of image in a category, we restrict the number of new image pairs to 100 per object category. This prevents over population of image pairs from a single category which will result in the transformation model biased towards transformations from that specific object class. The given pairwise combinations is done on top of $D_l$ resulting in 1800 additional image pairs, referred to as $D_{ul}$. Also, as the pairs are formed by pairwise combinations between images from labeled image pairs, we still have the semantic key-point locations as ground-truth for each image in the unlabeled image pair. The viewpoint variation is much higher in $D_{ul}$ as compared to $D_l$. This is shown with example image pairs in Figure 1 of Supplementary.

The combined set $D = D_l \cup D_{ul}$ forms our weakly-labeled training dataset. It is also ensured all test or validation image pairs are removed from $D$. In addition, direct flips of around 100 test image pairs are present in the training set. Although existing methods ignore this bias, in our case enforcing cyclic consistency or forward-backward loss in Section \ref{leobj} will essentially imply training the network on the test set. To avoid this bias, we further remove training samples that are flips of test pairs.

\noindent\textbf{Evaluation criteria.}
In line with previous work \cite{proposal_flow},\cite{ignacio_cvpr'18}, the proposed approach is evaluated by measuring the probability of correctly transferred key-points (PCK). This is given by the number of source key-points whose projections lie within a given threshold to the corresponding target points. The key-points are normalized by respective image width and height to the range [0,1]. The distance threshold for PCK is set to 0.1 for all experiments.

\subsection{Baselines and Methods}
We compare our proposed method with both CNN and traditional descriptor based methods. LOM\cite{proposal_flow}, and, OADSC\cite{yang2017object} use HOG descriptors to generate a dense correspondence map. SCNet \cite{scnet} and its variants use off-the-shelf region proposal methods to pool CNN descriptors, followed by geometrically constrained matching. Geometric transformation networks \cite{ignacio_cvpr'17}, and, \cite{ignacio_cvpr'18}, referred to as CNNGeo and CNNGeo2 respectively, directly regress transformation parameters that define the semantic mapping. CNNGeo is trained in a self-supervised manner using synthetic transformations, while, CNNGeo2 is trained on real image pairs using geometrically consistent feature correlation as a loss function.

As our proposed method is based on CNNGeo, we follow a similar abbreviation for our proposed methods. CNNGeo trained using Equations \ref{nn} and \ref{CD} are termed CNNGeo-NN and CNNGeo-CD respectively. Cyclic consistency loss with the nearest neighbor is termed as CNNGeo-NN-Cyc, while with Chamfer distance is abbreviated as CNNGeo-CD-Cyc. In addition, we trained CNNGeo2 with the image pairs in the dataset $D$, referred to as CNNGeo2*.

\subsection{Implementation details}

\noindent\textbf{Network Architecture.} We use the same network architecture as used in the baseline models, CNNGeo \cite{ignacio_cvpr'17}, and, CNNGeo2 \cite{ignacio_cvpr'18}. The network consists of a feature extraction layer which is a ResNet-101 \cite{resnet} architecture truncated at the \emph{conv4-23} layer. This is followed by a feature correlation layer and a series of convolutional layers. The final layer is a fully connected layer that outputs the parameters of the transformation model. The geometric model used is thin-plate spline which has 18 parameters.


\noindent\textbf{Training details.} We initialize the network parameters using CNNGeo. The proposed methods:CNNGeo-NN, CNNGeo-CD, CNNGeo-NN-Cyc and CNNGeo-CD-Cyc share the same training details. They are trained using the image pairs from the training set, $D$, described in Section \ref{sec:dataset}. The training images are resized to 240 $\times$ 240 resolution before feed-forwarding through the network. The network is implemented in Pytorch \cite{pytorch}, and, back-propagation is done using Adam \cite{adam}. Batch size is set to 16 with a learning rate of $5.10^{-6}$.

\subsection{Results}

We evaluated the baselines and existing methods on the PF-PASCAL test set and present our results in Table \ref{tab:results}. Overall, the proposed weakly-supervised approach outperforms the existing methods and the baseline geometric transformation models. The comparison with SCNet is not direct as we use ResNet-101 architecture which learns powerful representation than VGG-16 used by SCNet. The comparison with models CNNGeo and CNNGeo2 are also not direct, as we use additional supervision in the form of semantic key-points, but, do not use the correspondence ground-truth which is the end task. However, SCNet and its variants do use correspondence information in a weakly-supervised sense.

The results in Table \ref{tab:results} show that the model CNNGeo achieves better semantic matching performance when trained with the proposed loss functions. All the proposed methods (CNNGeo-*) outperform the existing methods across majority of the object categories. It is also observed that the cyclic consistency has comparable performance when combined with the nearest-neighbor and chamfer distance loss functions as can be observed from the performance of CNNGeo-NN-Cyc and CNNGeo-CD-Cyc. Results also show that clear improvement is obtained by methods that use cyclic consistency.

\begin{figure*}[t!]
		\centering
        \includegraphics[width=0.7\textwidth,height=0.4\textheight]{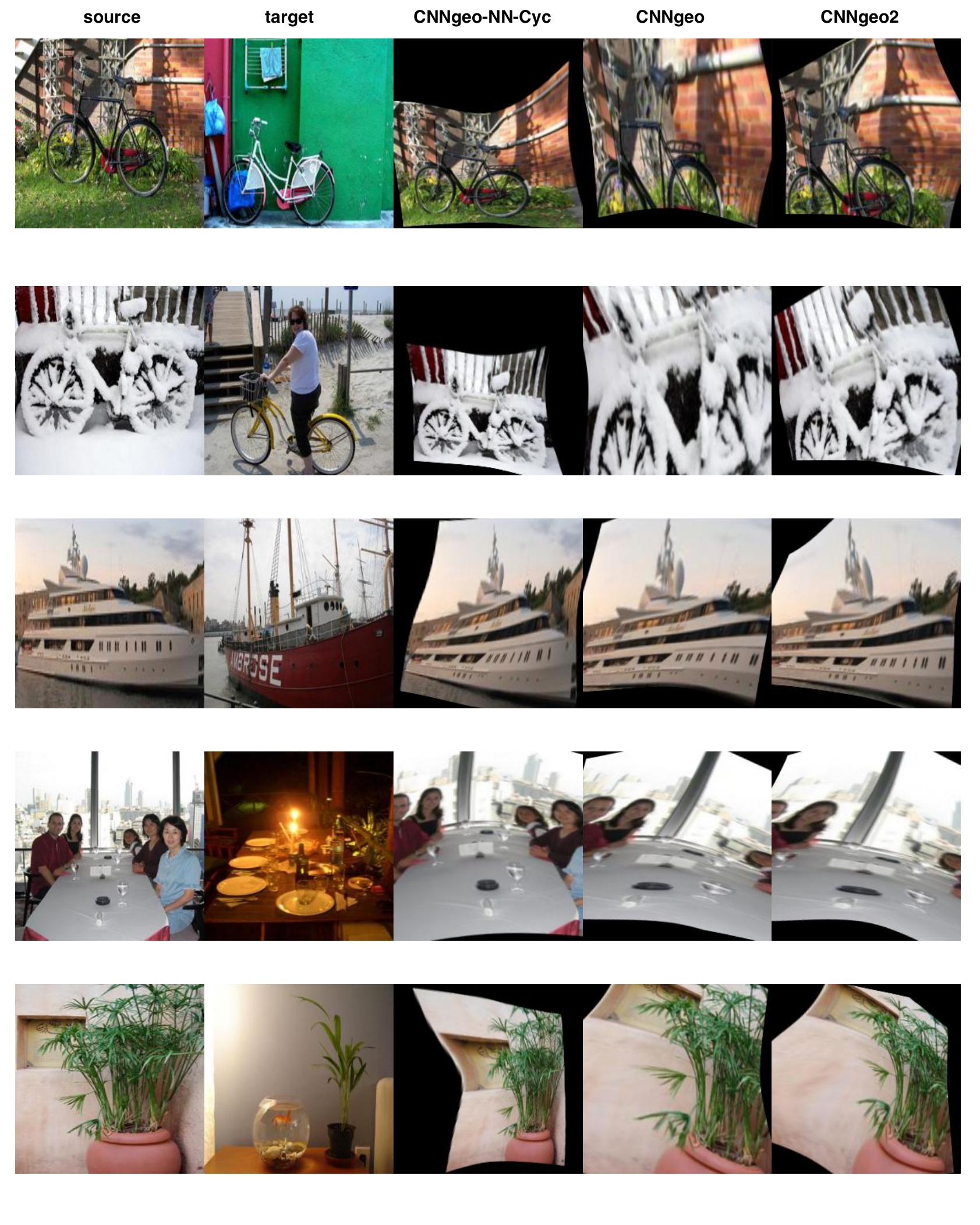}
        \caption{\textbf{Performance study.}. Each row in the Figure shows a different object. The samples are taken from the test set from PF-PASCAL dataset. Each example row is divided into 5 columns: \emph{i)}source image, \emph{ii)}target image, \emph{iii)}source image warped using the proposed CNNGeo-NN-Cyc,\emph{iv)}source image warped using baseline CNNGeo, and, \emph{v)}source image warped using CNNGeo2. \label{fig:res}}
\end{figure*}

\subsection{Generalization Performance}
Results demonstrate that with little extra supervision, the baseline method \cite{ignacio_cvpr'17} can benefit using the proposed approach. However, the extra supervision should not result in the the learnt geometric mapping over-fitting the source dataset. We therefore evaluate the generalization performance of the proposed approach on Caltech-101 \cite{caltech} and TSS \cite{taniai} datasets without further fine-tuning.

Caltech-101 dataset consists of 1515 image pairs from 101 object categories, and, was first used in \cite{kim2013deformable} for the task of semantic alignment. Semantic matching performance is evaluated using the following metrics: \emph{i)} the label transfer accuracy (LT-ACC); \emph{ii} the intersection-over-union (IoU), and, \emph{iii)} the object localization
error (LOC-ERR). On the other hand, TSS contains 400 image pairs divided into three categories:FG3DCar, PASCAL and JODS. Evaluation criteria is dense PCK computed over the foreground object. The threshold for PCK was set to 0.05 following evaluation settings in \cite{ignacio_cvpr'18}.

The proposed approach achieves state-of-the-art on Caltech-101 dataset under the LT-ACC metric as shown in Table \ref{generalize:caltech}. Overall, our approach CNNGeo-NN-Cyc consistently improves over the baseline CNNGeo as shown by gray highlights in Table \ref{generalize:caltech}.

From the results on TSS dataset as shown in Table \ref{generalize:tss}, CNNGeo-NN-Cyc outperforms CNNGeo on PASCAL category by 2 percentage points, while performs comparably in other categories. 

This shows that the proposed approach utilizes the additional supervision of semantic key-points well and generalizes to new datasets. Comparable performance is obtained by CNNGeo2 on both datasets. One reason could be the bound on representational capacity of the base network CNNGeo\footnote{Both CNNGeo2 and CNNGeo-NN-Cyc are trained by fine-tuning CNNGeo}. 

\begin{table}[]
\centering
\setlength\tabcolsep{1.5pt}
\scalebox{0.9}{
\begin{tabular}{|llll|}
\hline
                  & LT-ACC & IoU & LOC-ERR  \\ \hline \hline
LOM        & 0.78  & 0.50  & \hspace{0.5cm}0.26  \\ 
HOG+OADSC & 0.81 & 0.55 & \hspace{0.5cm}\textbf{0.19} \\
SCNet-A           & 0.78  & 0.50  & \hspace{0.5cm}0.28  \\ 
SCNet-AG          & 0.78  & 0.50  & \hspace{0.5cm}0.27  \\ 
SCNet-AG+         & 0.79  & 0.51  & \hspace{0.5cm}0.25  \\ 
CNNGeo            & 0.83  & 0.61  & \hspace{0.5cm}0.25   \\ 
CNNGeo2 & 0.85 & \textbf{0.63} & \hspace{0.5cm}0.24 \\
CNNGeo-NN-Cyc & \hspace{-0.07cm}\colorbox{lightgray}{\textbf{0.86}} & \hspace{-0.07cm}\colorbox{lightgray}{0.62} & \hspace{0.37cm}\colorbox{lightgray}{0.22} \\ \hline
\end{tabular}}
\caption{\textbf{Generalization performance on Caltech-101}.\label{generalize:caltech}}
\end{table}

\begin{table}[]
\centering
\setlength\tabcolsep{1.5pt}
\scalebox{0.9}{
\begin{tabular}{|llll|}
\hline
                  & FG3D & PASC & JODS  \\ \hline \hline
LOM\cite{proposal_flow}        & 0.786  & 0.531  & 0.653  \\ 
HOG+OADSC\cite{yang2017object} & 0.875 & \textbf{0.729} & 0.708 \\
CNNGeo  & 0.906  & 0.563  &\textbf{0.764}   \\ 
CNNGeo2 & \textbf{0.907} & 0.565 & \textbf{0.764} \\
CNNGeo-NN-Cyc & 0.903 & 0.593 & 0.755 \\ \hline
\end{tabular}}
\caption{\textbf{Generalization performance on TSS.} Error metric is PCK computed at threshold, $\alpha$ = 0.05. The generalization performance of the proposed method is comparable to baseline models as expected. \label{generalize:tss}}
\end{table}

\begin{table*}[]
\centering
\setlength\tabcolsep{1.5pt}
\scalebox{0.85}{
\begin{tabular}{|l|l|llllllllllllllllllll|l|}
\hline
             & Dataset     & aero  & bike  & bird  & boat  & bottle & bus   & car   & cat   & chair & cow   & table & dog   & horse & mbike & person & plant & sheep & sofa  & train & tv    & mean  \\ \hline
\hline
CNNGeo-NN    & $D_l$  & 85.5 & 88.4 & 85.3 & 76.4 & 68.8     & 95.4  & 93.3 & 84.6 & 50.6    & 91.7  & 38.0        & 72.9 & 62.9 & 80.2     & 73.7  & 72.9          & 100.0   & 66.3  & 74.3 & 65.6 & 78.4 \\ 
CNNGeo-NN    & $D$  & 86.1 & 87.1 & 79.7 & 70.8 & 70.3    & 98.1  & 93.0 & 74.2 & 54.5    & 91.7  & 32.8        & 70.3 & 66.2 & 76.2     & 69.4  & 65.2          & 80.0   & 50.3  & 77.8 & 78.9 & 76.7 \\ 
CNNGeo-NN-Cyc     & $D$ & 89.9 & 90.2 & 88.6 & 70.8 & 89.1   & 97.5 & 94.9  & 83.1 & 68.8 & 93.8 & 64.1          & 83.1 & 83.5 & 85.3     & 77.7  & 87.1        & 80.0   & 77.9 & 87.0    & 83.9 & 85.7 \\ \hline
\end{tabular}}
\caption{\textbf{Per class PCK on PF-PASCAL dataset}. PCK threshold, $\alpha$ = 0.1. An ablation study on the impact of cyclic consistency under viewpoint variation. Results show that addition of image pairs from $D_{ul}$ ($D = D_l \cup D_{ul}$) results in performance drop of simple nearest-neighbor based model, CNNGeo-NN. However, by adding cyclic consistency constraint, the resulting model CNNGeo-NN-Cyc achieves better performance  .\label{abl-1}}
\end{table*}

\begin{table*}[]
\centering
\setlength\tabcolsep{1.5pt}
\scalebox{0.85}{
\begin{tabular}{|l|l|llllllllllllllllllll|l|}
\hline 
& Dataset     & aero  & bike  & bird  & boat  & bottle & bus   & car   & cat   & chair & cow   & table & dog   & horse & mbike & person & plant & sheep & sofa  & train & tv    & mean \\ \hline
\hline
CNNGeo-NN-Cyc      & $D_l$    & 83.9 & 89.8 & 88.8 & 77.8 & 76.6  & 97.6 & 93.4 & 82.9 & 55.4 & 91.7  & 38.6        & 77.6 & 73.1 & 81.5     & 76.9  & 79.5       & 100.0    & 69.5 & 79.5 & 69.4 & 80.9 \\ 
CNNGeo-NN-Cyc & $D$ & 89.9 & 90.2 & 88.6 & 70.8 & 89.1   & 97.5 & 94.9  & 83.1 & 68.8 & 93.8 & 64.1          & 83.1 & 83.5 & 85.3     & 77.7  & 87.1        & 80.0   & 77.9 & 87.0    & 83.9 & 85.7 \\ 
CNNGeo2*  & $D$ & 84.3  & 87.7    & 81.4  & 54.2  & 64.1   & 90.5  & 92.8  & 86.2  & 46.6  & 91.7  & 31.3         & 76.6  & 70.5    & 75.9        & 70.1   & 78.6        & 100.0    & 57.9  & 61.3  & 63.9  & 75.6  \\\hline
\end{tabular}}
\caption{\textbf{Per class PCK on PF-PASCAL dataset}. PCK threshold, $\alpha$ = 0.1. Assessment of unlabeled samples $D_{ul}$ in semantic matching performance. The results show that addition of image pairs from $D_{ul}$ ($D = D_l \cup D_{ul}$) leads to the the proposed method, CNNGeo-NN-Cyc learning a better geometric mapping.  In comparison, baseline model, CNNGeo2* performs comparably to CNNGeo2 (c.f. Table \ref{tab:results}).\label{abl-2}}
\end{table*}

\begin{figure*}[t!]
		\centering
        \includegraphics[width=0.7\textwidth,height=0.4\textheight]{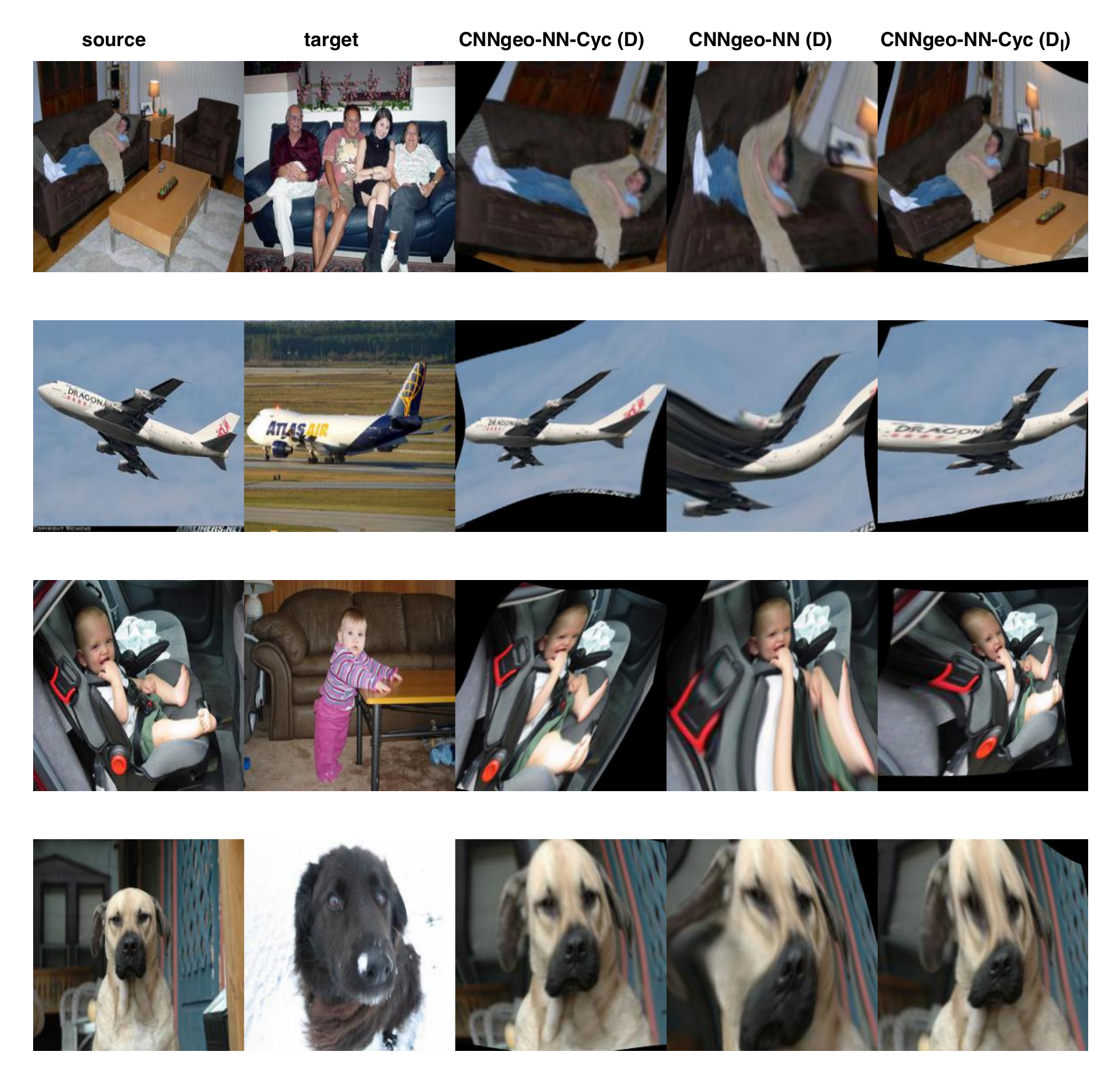}
        \caption{\textbf{Performance study.}. Each row in the Figure shows a different object. The samples are taken from the training data $D$. Each example row is divided into 5 columns: \emph{i)}source image, \emph{ii)}target image, \emph{iii)}source image warped using the proposed CNNGeo-NN-Cyc trained on dataset $D$,\emph{iv)}source image warped using CNNGeo-NN trained on $D$, and, \emph{v)}source image warped using CNNGeo-NN-Cyc trained on $D_l$. \label{fig:abl}}
\end{figure*}    

\subsection{Ablation Study}
\label{ablation}
In this section, we study two important aspects of our experiments. Firstly, we try to understand the interplay between the nearest-neighbor and cyclic consistency loss functions. And, secondly, we study the effect of unlabeled data in the performance.

The training samples in $D_l$ consist mostly of image pairs with similar viewpoint. On the other hand, as mentioned in Section \ref{sec:dataset}, $D_{ul}$ has a larger viewpoint variation. We train CNNGeo-NN on both $D_l$ and $D$ and present semantic matching performance on PF-PASCAL test set in Table \ref{abl-1}. Results show that addition of unlabeled pairs leads to a decrease in performance. This can be attributed to the viewpoint variation resulting in incorrect assignment of nearest-neighbor correspondence between source and target key-points. But, by adding the cyclic consistency constraint, the performance improves by a big margin demonstrating the importance of the proposed constraint in semantic point-set registration.

To study the impact of unlabeled data, $D_{ul}$ in learning a better geometric mapping, we train CNNGeo-NN-Cyc on both $D_l$ and $D$. From Table \ref{abl-2}, it can be observed that utilizing more data brings significant improvement in semantic matching performance on PF-PASCAL test set. We also trained baseline model, CNNGeo2 on $D$ training set (CNNGeo2* in Table \ref{abl-2}). Performance is comparable to CNNGeo2, showing that the method is not able to gain improvement in semantic matching performance by utilizing additional training samples. 

\begin{figure}[t!]
		\centering
        \includegraphics[width=0.45\textwidth]{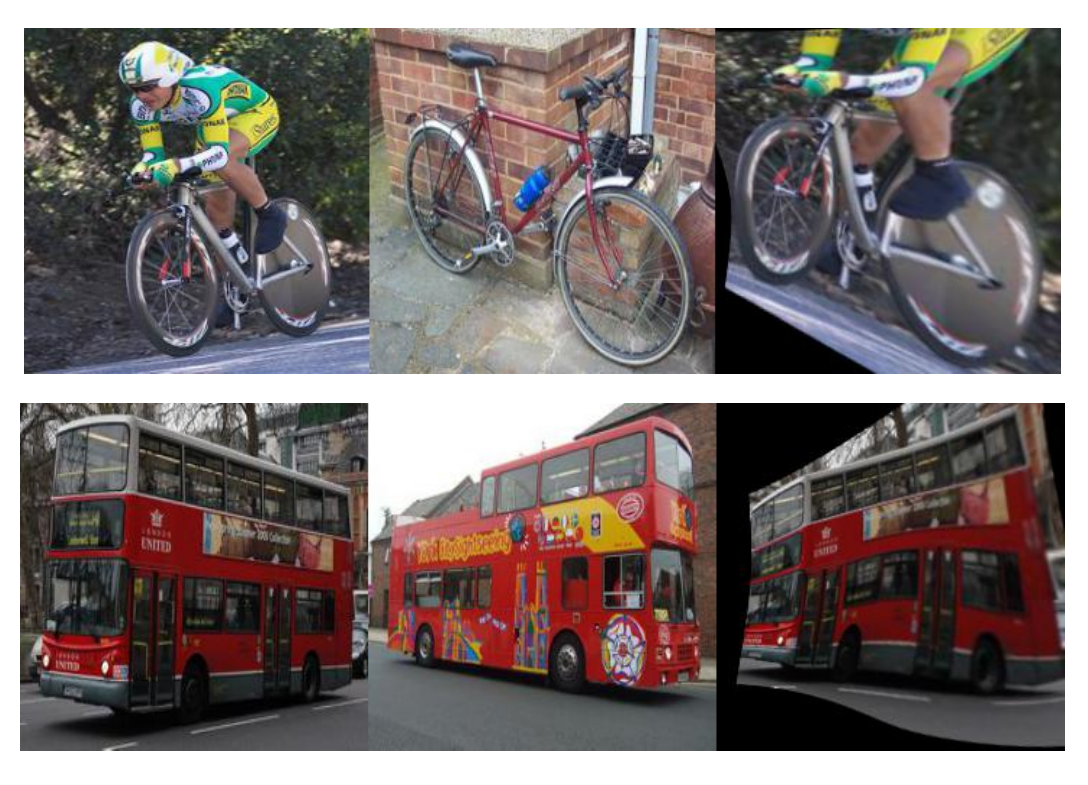}
        \caption{\textbf{Challenging cases.} \label{fig:fail}}
\end{figure}    

\subsection{Qualitative Results}
Besides the quantitative results reported in the previous sections, we also present a qualitative analysis of the experiments. Figures \ref{fig:res}, and, \ref{fig:abl} shows a pair of source-target image pairs of different objects (arranged row-wise), and the performance of different methods in aligning the source onto the target image using the estimated transformations. 

In Figure \ref{fig:res}, we show qualitative comparison of the warped source image by the baseline models, CNNGeo, CNNGeo2, and our best performing method, CNNGeo-NN-Cyc (c.f. Table \ref{tab:results}) on the test set of PF-PASCAL dataset. Results clearly demonstrate that our proposed method produces higher quality alignment than the baseline models.  

In Section \ref{ablation}, we made the observation that the combination of cyclic and nearest-neighbor constraint is important in learning to align semantically related images. Here we provide qualitative proof in Figure \ref{fig:abl} by comparing the alignment quality of warped source image using CNNGeo-NN-Cyc (column 3 in Figure), and, CNNGeo-NN (column 4). The samples shown in the Figure come from the set $D_{ul}$. The results show that CNNGeo-NN is not able to learn proprer semantic mapping purely based on nearest-neighbor constraint. Instead, by adding the cyclic consistency constraint, CNNGeo-NN-Cyc achieves much better semantic alignment.

We also show qualitative semantic alignment of CNNGeo-NN-Cyc trained on $D_l$ (5th column in Figure \ref{fig:abl}). The results show that the method generalizes quite well inspite of not seeing the samples during training. But, the performance is still behind the same method when trained using the full training set, $D$. This assessment and the results in Table \ref{abl-2} show that our proposed method leverages the additional training samples from $D_{ul}$ to achieve better semantic matching performance.

Despite achieving significant improvement over the baseline models, our proposed method (CNNGeo-NN-Cyc) still cannot solve certain challenging cases as shown in Figure \ref{fig:fail}. The samples are shown from the training set, $D$. However, the alignment quality is still acceptable given that only location information of the key-points was used as a weak supervision.

\section{Conclusion}
We presented a loss function for training a neural network to predict the transformation parameters for aligning an image pair. The loss function is based on nearest-neighbor cyclic consistency and only requires weak supervision in the form of overlapping set of key-points per image in a given image pair. Results demonstrate that our proposed approach outperforms the baseline models. Although our method uses additional supervision, it is still weakly supervised like the baseline models as no correspondence information is used during training.

In addition, we show that our proposed approach generalizes equally well as the baseline models to previously unseen data. Both quantitative and qualitative analysis is reported to demonstrate that the combination of cyclic and nearest-neighbor constraints is important in learning to align semantically related images.

\section*{Acknowledgments}
We acknowledge the computational resources provided by Aalto Science IT project and CSC servers, Finland.

{\small
\bibliographystyle{ieee}
\bibliography{egbib}
}

\IfFileExists{./supp.pdf}{
\clearpage
\includepdf[pages={1}]{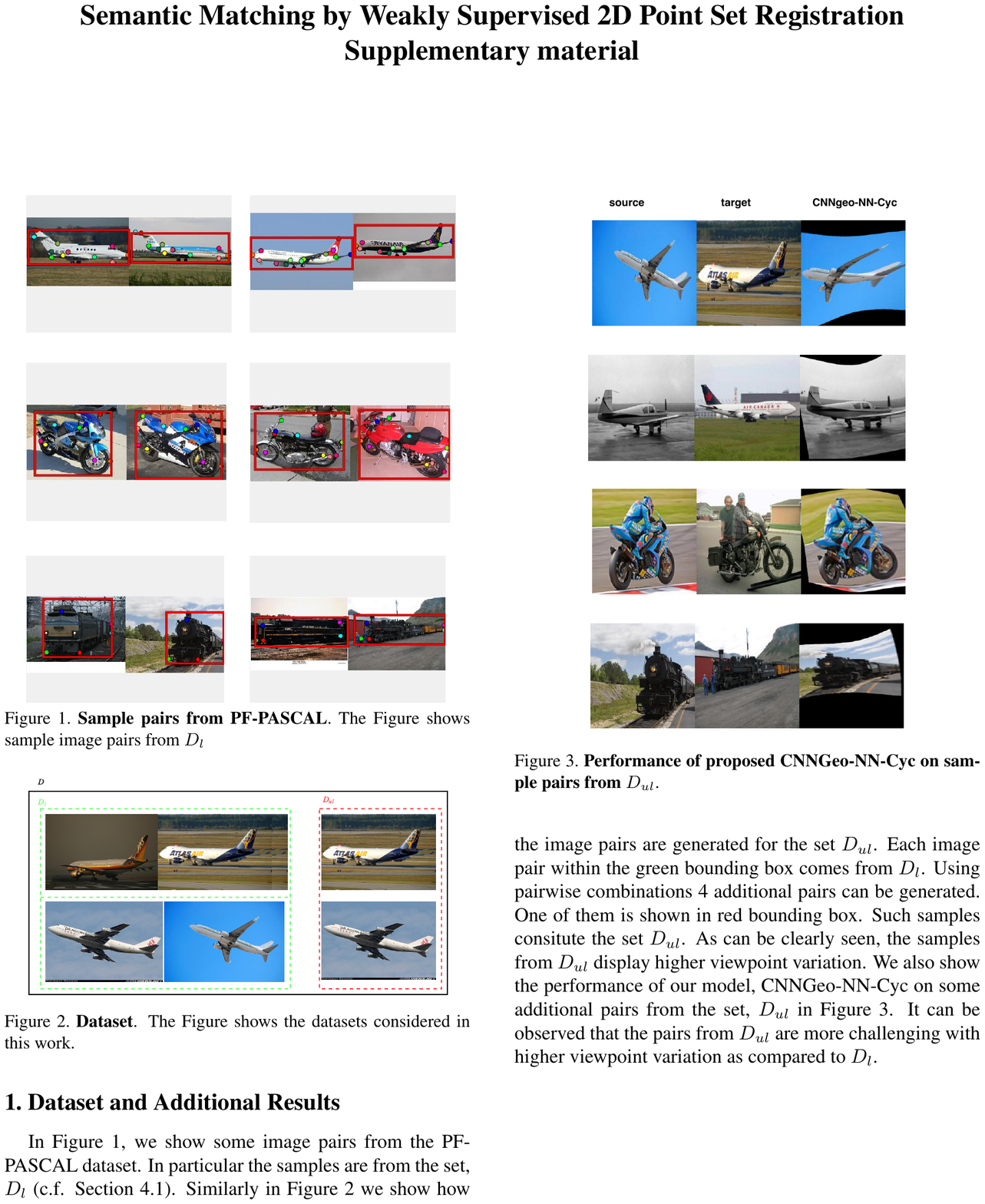}
}

\end{document}